\documentclass[runningheads]{llncs}
\usepackage{mathtools}
\usepackage{graphicx}
\usepackage{cite}
\usepackage{makecell}
\usepackage{multirow}
\usepackage{hyperref}
\usepackage{xcolor}
\renewcommand\UrlFont{\color{blue}\rmfamily}

\begin{document}
%
\title{AMINN: Autoencoder-based Multiple Instance Neural Network Improves Outcome Prediction in Multifocal Liver Metastases}
\titlerunning{AMINN: Autoencoder-based Multiple Instance Neural Network}
\author{Jianan Chen 
\inst{1,3}\and  
Helen M. C. Cheung\inst{2,3}\and
Laurent Milot\inst{4}\and
Anne L. Martel\inst{1,3}
}
\authorrunning{J. Chen et al.}

\institute{
Department of Medical Biophysics, University of Toronto, Toronto, ON, CA  
\\
\email{geoff.chen@mail.utoronto.ca} \and
Department of Medical Imaging, University of Toronto, Toronto, ON, CA \and
Sunnybrook Research Institute, Toronto, ON, CA\and
Centre Hospitalier Universitaire de Lyon, Lyon, FR}
%
%
%
%
\maketitle              

\begin{abstract}
Colorectal cancer is one of the most common and lethal cancers and colorectal cancer liver metastases (CRLM) is the major cause of death in patients with colorectal cancer. Multifocality occurs frequently in CRLM, but is relatively unexplored in CRLM outcome prediction. Most existing clinical and imaging biomarkers do not take the imaging features of all multifocal lesions into account. In this paper, we present an end-to-end autoencoder-based multiple instance neural network (AMINN) for the prediction of survival outcomes in multifocal CRLM patients using radiomic features extracted from contrast-enhanced MRIs. Specifically, we jointly train an autoencoder to reconstruct input features and a multiple instance network to make predictions by aggregating information from all tumour lesions of a patient. Also, we incorporate a two-step normalization technique to improve the training of deep neural networks, built on the observation that the distributions of radiomic features are almost always severely skewed. Experimental results empirically validated our hypothesis that incorporating imaging features of all lesions improves outcome prediction for multifocal cancer. The proposed AMINN framework achieved an area under the ROC curve (AUC) of 0.70, which is 11.4\% higher than the best baseline method. A risk score based on the outputs of AMINN achieved superior prediction in our multifocal CRLM cohort. The effectiveness of incorporating all lesions and applying two-step normalization is demonstrated by a series of ablation studies. A Keras implementation of AMINN is released\footnote{\UrlFont{https://github.com/martellab-sri/AMINN}}.

\end{abstract}
\section{Introduction}
Colorectal cancer (CRC) is the fourth most common non-skin cancer and the second leading cause of cancer deaths in the United States \cite{cancerstatistics}. 50-70\% of CRC patients eventually develop liver metastases, which become the predominant cause of death \cite{livermetsinstancerate}. Hepatic resection is the only potential cure for colorectal cancer liver metastases (CRLM), providing up to 58\% 5-year overall survival, compared to 11\% in patients with chemotherapy alone \cite{ferrarotto2011durable,58percentsurvivalrate}. Stratifying CRLM patients can lead to better treatment selection and improvements in prognosis.

 Main-stream clinical risk scores utilize clinical factors for CRLM (number of metastases$>$1, lymph node status, \textit{etc.}) for prognosis and treatment planning , but none are strong predictors of long-term survival \cite{roberts2014performance}. 
 A variety of medical imaging biomarkers have also been developed based on texture and intensity features extracted from the tumors and/or liver tissue in routinely collected Magnetic Resonance Imaging (MRI) to predict CRLM patient outcome\cite{nakai2020mri, chen2019unsupervised,cheung2018late}. 

Most medical-imaging-based prognostic models (not limited to CRLM) focus solely on the largest lesion when dealing with multifocality. Incorporating features from multiple lesions in modeling is difficult because it's unknown how each tumor contributes to patient outcome. Although 90\% of CRLM patients present with multifocal metastases, there is currently no medical-imaging based prognostic biomarker designed specifically for multifocal CRLM \cite{cancerstatistics}. In this paper, we address this gap with a radiomics-based multiple instance learning framework.

\begin{figure}
    \centering
    \includegraphics[width=\textwidth]{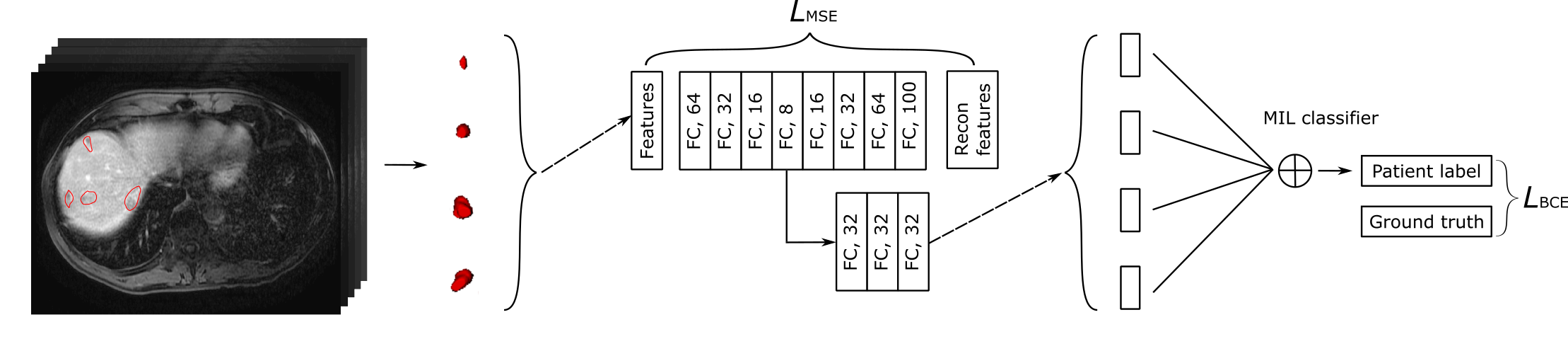}
    \caption{An overview of the proposed autoencoder-based multiple instance learning network. The network structure between the curly brackets is shared for each tumor of the same patient.}
    \label{fig:aminn}
\end{figure}

We hypothesize that using features from all lesions of multifocal cancers improves outcome prediction. We propose an end-to-end autoencoder-based multiple instance neural network (AMINN) that predicts multifocal CRLM patient outcome based on radiomic features extracted from contrast-enhanced 3D MRIs (\textbf{Fig. \ref{fig:aminn}}). This model is jointly trained to reconstruct input radiomic features with an autoencoder branch and to make predictions with a multiple instance branch. This two-branch structure encourages the learning of prognosis-relevant representations of tumors. The multiple instance neural network aggregates the representations of all tumors of a patient for survival prediction. In addition, we propose a feature transformation technique tailored to radiomic features to suit the need of deep learning models. Ablation experiments were performed to show the effect of all the components of our framework.

\section{Methods}
\textbf{Multiple instance learning (MIL):} MIL is a class of machine learning algorithms that learn from a bag of instances, where labels are available at the bag-level but not at instance level \cite{maron1998framework}. MIL has been widely adopted in medical image and video analysis as it performs well in weakly-supervised situations \cite{quellec2017multiple}. In most cases, detection or prediction using medical images is modeled as a MIL problem where the whole image (or 3d scan) is a bag and image patches (or cubes) cropped from the image are instances. Our work differs from previous MIL studies in treating multifocal tumors as instances. This allows us to generate more accurate instance representations (as image patches can sometimes contain redundant or irrelevant information) and also simplify the MIL problem by reducing the number of instances.

We formulated the task of patient outcome prediction as a MIL problem as follows: 
\begin{equation}
p_{i} = \mathcal{M} \left ( f \left ( \Phi\left ( x_{ij|j=1...k} \right ) \right )\right ) \label{eq1}
\end{equation}
where $p_{i}$ is the probability of event $y_{i} \in \{0,1\}$ (3-year overall survival) for patient $i$ and $X_{i} = \{x_{i1},...,x_{ik}\}$ is a bag of instances representing a varying number of $k$ tumors in patient $i$. $\Phi$ represents the encoding function which maps input features $x_{ij}$ to latent representations $h_{ij}$. $f$ represents the transformation function parameterized by the neural network for extracting information from instance-level representations and $\mathcal{M}$ is a permutation invariant multiple instance pooling function that aggregates instance representations to obtain bag representations and bag labels.

\textbf{Radiomics features:} Radiomics is an emerging field where mathematical summarizations of tumors are extracted from medical images for use in modeling\cite{gillies2016radiomics}. Radiomic approaches have shown promising results in various tasks and imaging modalities. For the analysis of multifocal cancers, radiomic models usually use features from the largest tumor, resulting in the loss of information \cite{yip2016applications}. Deep learning-based workflows have been investigated as an alternative to conventional radiomics, however, building a generalizable deep learning-based  model requires huge datasets that are currently impractical in medical imaging and radiomics is still the generally preferred approach  \cite{afshar2019handcrafted}. 

We extract radiomic features from preoperative 10-minute delayed T1 MRI scans using pyradiomics (v2.0.0)\cite{van2017computational}. The 3D MRI scans were resampled to [1.5, 1.5, 1.5] isotropic spacing using B-spline interpolation. Images were normalized using Z-score normalization with 99th percentile suppression, rescaled to range [0, 100] and discretized with a bin size of 5. Tumors are delineated slice by slice by a radiologist with 6 years of experience in abdominal imaging. For each tumor lesion, we extract 959 features (including 99 original features, 172 Laplacian of Gaussian filtered features and 688 wavelet features) from three categories to mathematically summarize its intensity, shape and texture properties. However, we only kept the 99 original features as the models trained with original features performed the best. A list of the final features is included in \textbf{Table S1}.

\begin{figure}
    \centering
    \includegraphics[width=\textwidth]{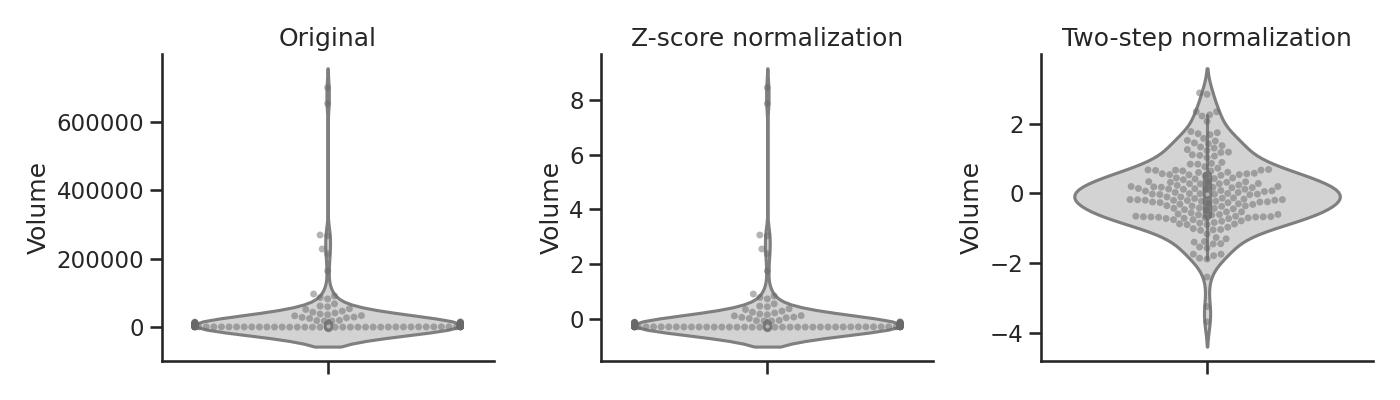}
    \caption{Volume of 181 lesions from 50 patients with Z-score normalization and two-step normalization.}
    \label{fig:normalization}
\end{figure}

\textbf{Feature normalization}: Normalization is a common technique in machine learning for bias correction and training acceleration. Ideally, for deep learning frameworks, input data should be scaled, centered at zero and relatively symmetrically distributed around the mean to alleviate oscillating gradients and avoid vanishing (due to packed points) or exploding (due to extreme values) gradients in gradient descent \cite{shi2000reducing}. We observe that the distribution of most of our radiomic features are severely skewed, calling for normalization. However, Z-score normalization, one of the most commonly-used techniques for radiomic feature normalization, is less informative for data that is not approximately normally distributed. As an example, after Z-score normalization of tumor volume, above 90\% of samples are packed in the range [-0.5, 0], while the largest lesions have Z-value up to 8 (\textbf{Figure \ref{fig:normalization}}). This is undesirable as tumor volume is correlated with many radiomic features and is an important predictor of patient outcome\cite{welch2019vulnerabilities}.

We utilize a simple two-step normalization algorithm that combines logarithmic and Z-score transformation to tackle this problem:
\begin{align}
l_{ij} &= \log(f_{ij}-2\mathrm{min}(F_{j}) + \mathrm{median}(F_{j}) ) \\ \label{eq2}
Z_{ij} &= \frac{l_{ij}-\mu}{\sigma}
\end{align}
where $f_{ij} \in F_j=\{f_{1j}, f_{2j}, ... f_{nj}\}$ is the $j$th input feature of the $i$th sample. A logarithmic transformation is first applied to reduce skewness. To handle negative and zero values, we incorporate a feature-specific constant, $-2\mathrm{min}(F_j) + \mathrm{median}(F_j)$.
It is a combination of a first $-\mathrm{min}(F_j)$ term, that ensures values are larger than or equal to zero, and a second term $\mathrm{median}(F_j)-\mathrm{min}(F_j)$, that aims to avoid log operations on extremely small numbers. This is followed by a standard Z-score transformation(\textbf{Fig. \ref{fig:normalization}}).


\textbf{Network structure}: The autoencoder applied to reduce the dimensionality of input radiomic features is composed of a four-layer encoder and a four-layer decoder. All layers are fully-connected layers with ReLU activation except the last layer, which uses sigmoid activation. While autoencoders can be employed to select radiomic features by learning their low-dimensional representations in an unsupervised manner \cite{chen2019unsupervised}, we wish to directly extract the most relevant latent representations for patient prognosis. Thus the bottleneck layer of the autoencoder is connected to the 
multiple instance prediction network as input in order to select latent features that are informative of patient outcome.

The MIL network consists of 3 fully connected layers followed by a MIL pooling layer that integrates instance representations to predict patient label. Each fully connected layer consists of 32 hidden units. Various pooling methods are tested with the same backbone network\cite{wang2018revisiting,zhou2019handbook}:

$$
\mathcal{M} = \left\{
    \begin{array}{ll}
        max: p = \underset{j=1,...k}{\mathrm{max}} {f_{j}} \\
        average: p = \frac{1}{k}\sum_{j=1}^{k}f_{j} \\
        lse: p=r\mathrm{log}\left [ \frac{1}{k} \sum_{j=1}^{k}\mathrm{exp}\left ( rf_{j}\right )\right ]\\
        att: p = \sum_{j=1}^{k}a_{j}f_{j}
    \end{array}
\right.
$$

The selection of MIL pooling functions usually depends on the specific task. Since it is unclear how multiple tumors collectively determine clinical outcome, in addition to \textit{max}, \textit{average} and \textit{lse} (log-sum-exp, a smooth approximation of \textit{max}), we also test attention-based pooling (\textit{att})\cite{ilse2018attention}. The \textit{att} model incorporates a trainable pooling module that learns the importance of each instance, and can be more flexible and interpretable compared to standard pooling functions \cite{zhou2019handbook}. 

\section{Experiments}
We used a retrospective cohort of 108 colorectal cancer liver metastases (CRLM) patients that were eligible for hepatic resection and were treated at our institution. Informed consent was waived by the institutional review board. All patients received Gadobutrol-enhanced MRI prior to hepatic resection. 50 patients had multifocal CRLM, and each had 2-17 lesions (mean 3.6), leading to a total of 181 lesions. Median patient overall survival was 30 months and was right-censored to obtain 3-year survival (19/50 events). The cohort has been described in detail in previous work \cite{cheung2018late}. Demographics of the multifocal cohort is available in \textbf{Table S2}.

We used the unifocal patient subset of our dataset as an independent cohort for hyper-parameter tuning using grid search in 10 repeated 3-fold cross validation (CV). Using the selected hyper-parameters, we trained AMINN on the multifocal patient subset over 10 repeated 3-fold CV with splits stratified by patient outcome, each in 100 epochs without dropout. The model was optimized using Adam optimization with an initial learning rate of 0.0001, $\beta_1=0.9$ and $\beta_2=0.999$, on a compound loss -- mean squared error loss for reconstruction of the autoencoder plus binary cross-entropy loss from the patient-level prediction:
  \begin{equation}
      \mathbf{Loss} = \mathbf{L}_{\mathrm{MSE}}(x_{recon}, x) + \alpha\mathbf{L}_{\mathrm{BCE}}(\hat{y},y)
      \label{eqloss}
  \end{equation}
where $\alpha$ is a hyperparameter for the trade-off between reconstruction error and prediction error. In our experiments, $\alpha=1$ provides robust results and given the small sample size of the dataset, we didn't attempt to tune this parameter.

 Models were evaluated using area under the ROC curve (AUC) with mean and 95\% confidence interval, and accuracy (ACC) with mean and standard deviation, estimated across 10 repeats of 3-fold cross validations for predicting 3-year survival. We chose AUC as the primary evaluation criteria because our dataset is slightly imbalanced. Commonly-used models in radiomic analysis: support vector machine (SVM), random forest (RF) and logistic regression (LR), combined with LASSO-selected features of the largest lesion were built as baselines \cite{tibshirani1996regression}. We also built an abridged version of AMINN trained only on the largest lesions of multifocal patients for comparison. Further, we performed a series of ablation experiments to evaluate the effect of each component in our network.

We predicted the outcome for each fold based on a model trained on the other two folds, and combined the outputs of each test fold to obtain the survival probability predictions for all patients. We derived a binary risk score by median dichotimizing the model outputs. We compared our binary score against other clinical and imaging biomarkers for CRLM, including the Fong risk score, a clinical risk score for treatment planning, the tumor burden score (TBS), a recent risk score calculated from tumor counts and size, and target tumor enhancement (TTE), a  biomarker specifically designed for 10-minutes delayed gadobutrol-enhanced MRIs \cite{fong1999clinical,tumorburdenscore,cheung2018late}. All risk scores were evaluated using concordance-index (c-index) and hazard ratio (HR) in univariate cox proportional hazard models. In addition, a multivariable cox model incorporating all risk scores as predictors was used to evaluate the adjusted prognostic value of our risk score.

\section{Results}

\begin{table}[]
    \centering
        \caption{Comparison of machine learning models for predicting multifocal CLRM patient outcome in 10 repeated runs of 3-fold cross validation. }
    \begin{tabular*}{\textwidth}{l@{\extracolsep{\fill}}llll}
    \hline
        \textbf{Model} & \textbf{Setting} & \textbf{AUC} (\textbf{95\%CI}) &  \textbf{Accuracy} \\ \hline 
         \multicolumn{4}{c}{\textit{Largest lesion}} \\ \hline
        SVM & Largest & 0.58 (0.54-0.62) & $0.62 \pm 0.08 $\\ \hline
        RF & Largest & 0.58 (0.55-0.62)& $0.60\pm 0.10$\\ \hline 
        LR & Largest & 0.58 (0.55-0.62) & $0.62\pm 0.07$\\ \hline 
        AMINN-unifocal &  Largest & 0.58 (0.57-0.59) & $0.61\pm 0.08 $\\ \hline
        \multicolumn{4}{c}{\textit{Multiple instances}} \\ \hline
        AMINN & Max & 0.67 (0.65-0.69) & $0.65 \pm 0.11$\\ \hline
        AMINN & LSE & 0.67 (0.66-0.68) & $0.67 \pm 0.09$\\ \hline
        AMINN & Average & 0.70 (0.67-0.73) & $0.68 \pm 0.09 $\\ \hline
        AMINN & Attention \cite{ilse2018attention}& 0.65 (0.62-0.68) & $0.65 \pm 0.10 $\\ \hline
    \end{tabular*}

    \label{tab:comparison}
\end{table}

The proposed autoencoder-based multiple instance neural network (AMINN) outperforms baseline machine learning algorithms by a large margin (\textbf{Table \ref{tab:comparison}}). AMINN with average pooling has the best performance in both area under the ROC curve (AUC) and accuracy (ACC). Compared to the best performing baseline method (logistic regression with LASSO feature selection based on the largest tumor), AMINN with average pooling achieves a 11.4\% increase in AUC and a 5.7\% increase in ACC. The unifocal version of AMINN, which has the same structure as AMINN but only uses features from the largest tumor, has similar performance as other baselines. Attention-based pooling doesn't show superior performance over other methods, suggesting that more data are required for the proposed framework to learn the lethality (weights) of each lesion. 

We then tested the performance gain from each component of AMINN, namely incorporating all lesions into outcome prediction (\textbf{multi}), two-step feature transformation (\textbf{log}), and feature reduction with an autoencoder (\textbf{ae}) (\textbf{Table \ref{tab:ablation}}). Incorporating all lesions consistently improved AUC, by 3.0\%, 4.1\%, 7.9\% and 8.3\% compared to AMINN with the baseline setting (using features from the largest lesion, z-score transformation and fully connected network), with \textbf{log}, with \textbf{ae}, and with \textbf{log+ae}, respectively. Notably, adding two-step normalization also boosts performance by 4.4\% to 6.5\% from AMINN using only Z-score normalization. In contrast, incorporating two-step normalization results in similar performance compared to Z-score normalization in baseline non-neural network models (for conciseness, only results for logistic regression are shown). This supports the hypothesis that two-step normalization improves the training of neural networks with radiomic features as inputs. The autoencoder component of AMINN only improves model performance when multiple lesions are considered. One possible explanation is that the autoencoder overfits quickly in the abridged models.

\begin{table}[]
    \centering
        \caption{Ablation studies with different components of AMINN. }
    \begin{tabular*}{\textwidth}{l@{\extracolsep{\fill}}lll}
    \hline
        \textbf{Model} & \textbf{AUC} (\textbf{95\%CI}) &  \textbf{Accuracy} \\ \hline 
        $Logistic Regression$ & 0.58 (0.54-0.62) & $0.62 \pm 0.07$ \\ \hline
        $Logistic Regression_{log}$ & 0.59 (0.54-0.63)& $0.63 \pm 0.09$ \\ \hline \hline
        $AMINN_{baseline}$  & 0.58 (0.57-0.59)& $0.61 \pm 0.08$\\ \hline
        $AMINN_{log}$  & 0.63 (0.63-0.64)& $0.61 \pm 0.05$\\ \hline 
        $AMINN_{ae}$  & 0.57 (0.56-0.59) & $0.57 \pm 0.08$\\ \hline 
        $AMINN_{log+ae}$ & 0.62 (0.59-0.64) & $0.58 \pm 0.07 $\\ \hline
        $AMINN_{multi}$ & 0.61 (0.59-0.63) & $0.62 \pm 0.11$\\ \hline
        $AMINN_{multi+log}$ & 0.68 (0.66-0.69) & $0.64 \pm 0.14$\\ \hline
        $AMINN_{multi+ae}$ & 0.65 (0.62-0.68) & $0.62 \pm 0.08$ \\ \hline
        $AMINN_{multi+log+ae}$  & 0.70 (0.67-0.73) & $0.68\pm0.09$\\ \hline
    \end{tabular*}
    \label{tab:ablation}
\end{table}
When compared to clinical and imaging biomarkers for CRLM, our risk score is the only one that achieved predictive value in univariate cox regression modeling for our cohort of multifocal patients (\textbf{Table \ref{tab:clinical}}), with c-index of 0.63 and HR of 2.88 (95\%CI: 1.12-7.74). In multivariable analysis where all four biomarkers are included as predictors, our risk score remained the only one showing predictive value. 

Interestingly, performance of all models improved when applied to the entire cohort of 58 unifocal and 50 multifocal cases (\textbf{Table 4}). We observe that although both TBS and the Fong score take into account the number of tumor lesions, they fail to further stratify within multifocal patients as all patients have elevated risk from multifocality. Similarly, although TTE incorporates intensity features of the two largest lesions when there are multiple tumours, it has less predictive value when all patients have two or more tumors. The gap in predictive power of these biomarkers between predicting on the full cohort and the multifocal subset highlights the need for developing multifocality-aware cancer prediction models.  

\begin{table}[]
    \centering
        \caption{Uni- and multi-variable cox regression models using different biomarkers on  multifocal cohort. Q-values derived from p-values  with Benjamini–Hochberg correction.}
    \begin{tabular*}{\textwidth}{|l@{\extracolsep{\fill}}|l|l|l|l|l|l|}
    \hline
     &\multicolumn{3}{l|}{\textbf{Univariate analysis}}&\multicolumn{3}{l|}{\textbf{Multivariable analysis}} \\ \hline
        \textbf{Method} & \textbf{HR (95\%CI)} & \textbf{C-index} & \textbf{Q-value} & \textbf{HR (95\%CI)}  &\textbf{C-index} & \textbf{Q-value}\\ \hline 
        Fong\cite{fong1999clinical}&1.44 (0.60-3.44) & 0.55& 0.41 & 1.19 (0.45-3.15)&\multirow{4}{*}{0.72} & 0.47\\ \cline{1-5} \cline{7-7}
        TBS \cite{tumorburdenscore}& 1.73 (0.77-3.88) & 0.58 & 0.18& 2.59 (0.96-7.01)&  & 0.06\\ \cline{1-5} \cline{7-7}
        TTE \cite{cheung2018late}& 2.29 (0.86-6.11) & 0.59 &  0.10 & 1.72 (0.63-4.76)&  & 0.29\\ \cline{1-5} \cline{7-7}
        AMINN &  2.88 (1.12-7.44) & 0.63 & 0.03* & 4.22 (1.38-12.88)&  & 0.01*\\ \hline

    \end{tabular*}

    \label{tab:clinical}
\end{table}

\begin{table}[]
    \centering
        \caption{Existing biomarkers have higher predictive value on the full cohort (unifocal subset plus multifocal subset)}
    \begin{tabular}{|c|c|c|c|}
    \hline
        \textbf{Method} & \textbf{HR (95\% CI)} & \textbf{C-index} & \textbf{p-value} \\ \hline 
        Fong score \cite{fong1999clinical}&1.96 (0.94-4.08) & 0.57& 0.07 \\ \hline
        TBS \cite{tumorburdenscore}& 1.72 (1.02-2.89) & 0.61 & 0.04\\ \hline 
        TTE \cite{cheung2018late}& 2.82 (1.38-5.77) & 0.65 &  $<$0.01\\ \hline 
    \end{tabular}
\end{table}


\section{Conclusion \& Discussion }
In conclusion, we propose an end-to-end autoencoder-based multiple instance neural network (AMINN) to predict outcomes of multifocal CRLM patients from radiomic features extracted from contrast-enhanced MRIs. By incorporating information from all lesions for patient outcome prediction, our model achieves a 11.4\% increase in AUC and 5.7\% increase in accuracy compared to commonly-used methods. The risk score built from the outputs of our network outperforms other clinical and imaging biomarkers and remains the only one with predictive value in our multifocal CRLM cohort. We also demonstrate the ability of our two-step normalization technique to improve the performance of our radiomic-feature-based neural network. 

The primary limitation of our study is its small sample size. Preoperative multifocal CRLM scans are difficult to acquire because most multifocal patients are not eligible for surgery and have poor survival \cite{livermetsinstancerate}. To the best of our knowledge, there is no public imaging dataset curated for outcome prediction of multifocal patients in any tumor type. One of the goals of this study is to draw attention to multifocal diseases in the medical image computing community.

Ideally, we would like to use an automatic segmentation model, for instance, the nnUNet \cite{isensee2021nnu} to perform tumor segmentation in order to avoid inter- / intra-observer variability and to automate the analysis pipeline. However, although we were able to obtain high accuracy in segmenting the whole liver using deep learning, we found that the results for liver tumor segmentation were unsatisfactory. We therefore opted to use manual segmentation by a radiologist in order to evaluate our methodology. The development of a more accurate detection and segmentation network for MRI liver lesions is the subject of future work. Similarly, we did not apply whole image-based deep learning because the features learned can be biased if the network cannot detect or segment lesions accurately.


\subsubsection{Acknowledgements}
The authors would like to thank The Natural Sciences and Engineering Research Council of Canada (NSERC) for funding, and acknowledge the contribution of Drs. Karanicolas, Law and Coburn in helping to create the patient cohort for this study.

\bibliographystyle{splncs04}
\bibliography{paper383.bbl}
\end{document}